\documentclass[sigconf, screen]{acmart}
\AtBeginDocument{%
  \providecommand\BibTeX{{%
    \normalfont B\kern-0.5em{\scshape i\kern-0.25em b}\kern-0.8em\TeX}}}

\setcopyright{acmcopyright}
\copyrightyear{2024}
\acmYear{2024}
\acmDOI{https://doi.org/10.1145/3610978.3640723}

\copyrightyear{2024}
\acmYear{2024}
\setcopyright{rightsretained}
\acmConference[HRI '24 Companion]{Companion of the 2024 ACM/IEEE International Conference on Human-Robot Interaction}{March 11--14, 2024}{Boulder, CO, USA}
\acmBooktitle{Companion of the 2024 ACM/IEEE International Conference on Human-Robot Interaction (HRI '24 Companion), March 11--14, 2024, Boulder, CO, USA}
\acmDOI{10.1145/3610978.3640723}
\acmISBN{979-8-4007-0323-2/24/03}

\makeatletter
\gdef\@copyrightpermission{
  \begin{minipage}{0.3\columnwidth}
   \href{https://creativecommons.org/licenses/by/4.0/}{\includegraphics[width=0.90\textwidth]{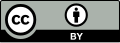}}
  \end{minipage}\hfill
  \begin{minipage}{0.7\columnwidth}
   \href{https://creativecommons.org/licenses/by/4.0/}{This work is licensed under a Creative Commons Attribution International 4.0 License.}
  \end{minipage}
  \vspace{5pt}
}
\makeatother

\begin{document}

\title{The Conversation is the Command: Interacting with Real-World Autonomous Robot Through Natural Language}

\author{Linus Nwankwo}
\email{linus.nwankwo@unileoben.ac.at}
\orcid{0000-0002-1767-2140}
\affiliation{%
  \institution{Chair of Cyber-Physical Systems, Montanuniversität}
  \streetaddress{Franz Josef-Straße 18, 8700 Leoben}
  \city{Leoben}
  \country{Austria}
  \postcode{8700}
}

\author{Elmar Rueckert}
\affiliation{%
  \institution{Chair of Cyber-Physical Systems, Montanuniversität}
  \streetaddress{Franz Josef-Straße 18, 8700 Leoben}
  \city{Leoben}
  \country{Austria}}
\orcid{0000-0003-1221-8253}


\renewcommand{\shortauthors}{Linus Nwankwo and Elmar Rueckert}
\begin{abstract}
In recent years, autonomous agents have surged in real-world environments such as our homes, offices, and public spaces. However, natural human-robot interaction remains a key challenge.
In this paper, we introduce an approach that synergistically exploits the capabilities of large language models (LLMs) and multimodal vision-language models (VLMs) to enable humans to interact naturally with autonomous robots through conversational dialogue. We leveraged the LLMs to decode the high-level natural language instructions from humans and abstract them into precise robot actionable commands or queries. Further, we utilised the VLMs to provide a visual and semantic understanding of the robot's task environment. Our results with \(99.13\%\) command recognition accuracy and \(97.96\%\) commands execution success show that our approach can enhance human-robot interaction in real-world applications.  The video demonstrations of this paper can be found at \url{https://osf.io/wzyf6} and the code is available at our repository\footnote{\url{https://github.com/LinusNEP/TCC_IRoNL.git}}.
\end{abstract}



\maketitle
\section{Introduction}
The exploration of human-robot interaction (HRI) \cite{hri}, \cite{hri1} and its advancement into real-world applications has been a topic of significant research over the past decades \cite{hri2}. Current approaches for controlling and interacting with autonomous robots in the real world have been dominated by complex teleoperation controllers \cite{teleop}, teach pendants \cite{kineTeach}, and rigid command protocols \cite{protocol}, where the robots execute predefined tasks based on specialized programming languages. As the challenges we present to these robots become more intricate and the environments they operate in grow more unpredictable \cite{linus}, there arises an unmistakable need for more natural and intuitive interaction mechanisms.

Prior works have seen a tilt towards techniques like reinforcement \cite{rl2} and imitation learning \cite{kineTeach}. By leveraging iterative learning and human demonstrations, these strategies have shown a capacity for fostering nuanced robot behaviours, as demonstrated in \cite{rl1}. However, the often computational burdens \cite{lynch2022interactive}, and the high costs associated with reward specification \cite{vlm}, task-specific training, or fine-tuning, commonly observed in the reinforcement and imitation learning frameworks, have limited the practical applicability of these methods, especially for simpler robotic tasks.

\begin{figure*}[htp]
    \centering   
    \includegraphics[scale=0.257]{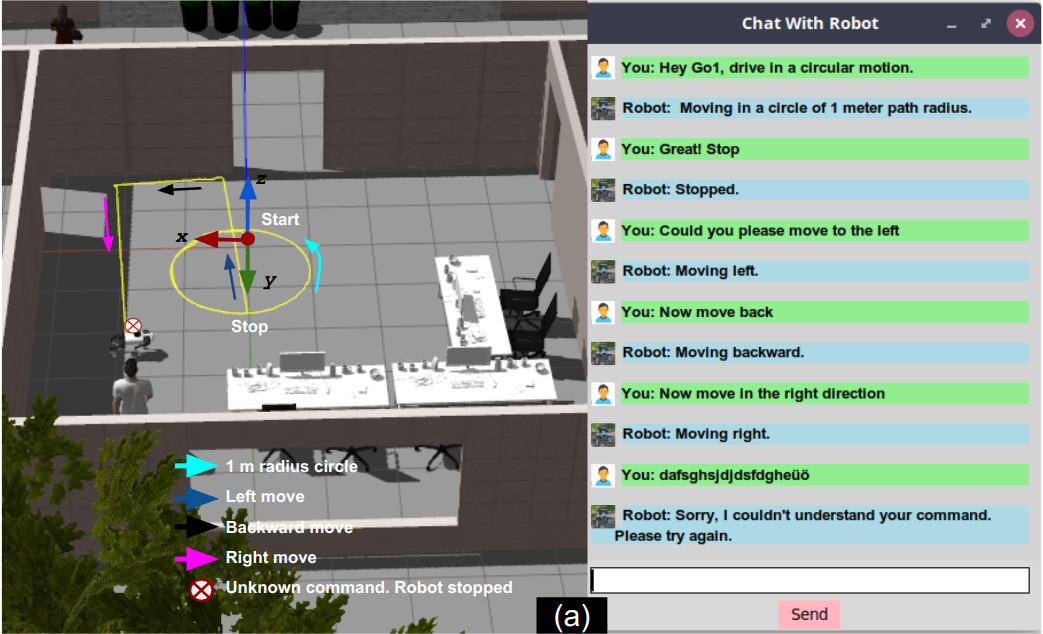}
    \includegraphics[scale=0.257]{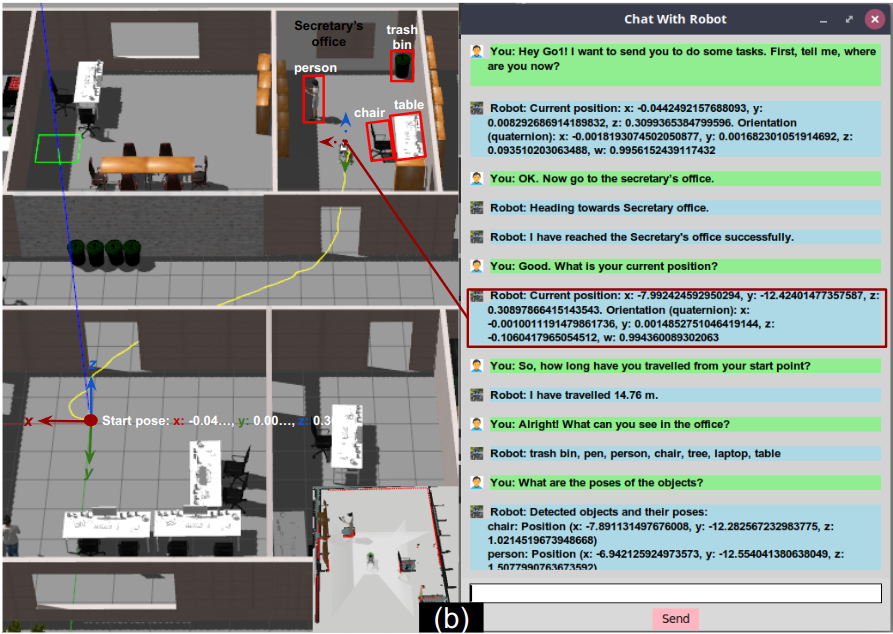}
    \caption{Example demonstration of our framework. We demonstrated these results in the real world as shown in the summary video at \url{https://osf.io/wzyf6}. In (a), our framework decodes the high-level instructions such as \textit{``move in a circular pattern''}, \textit{``move forward, go right, etc.''} from humans, and abstracts them to the robot's physical actions. In (b), we leveraged our framework for the robot's task environment understanding, information requests, and goal navigation.}
    \label{fig:exampleResults}
    \Description{These two figures (a \& b) are a simulation interface showing the interaction between a user and the robot within a virtual environment. The figure contains a Gazebo 3D world model of an office environment and a chat window for interacting with the robot. The Gazebo 3D world model of the office environment is marked with different objects such as a person, chairs, tables, and trash bins. On the right side of the figures is a chat window titled ``Chat With Robot'' where a conversation is taking place between a user and a robot. The user gives commands and asks questions, while the robot provides updates on its position and observations. The conversation includes coordinates, representing the robot's location in the environment, and descriptions of the robot's responses, e.g., travel distance and what it can see in the environment. The bottom part of the first figure and the top part of the second figure show more information about the commands and the objects the robot has detected, including their positions.}
\end{figure*}

Prompted by these challenges, we turned our focus to the recent advancement in large language models (LLMs) \cite{gpt2}, \cite{gpt3} and multi-modal vision-language models (VLMs) \cite{clip}, \cite{dalle} to foster an intuitive human-robot collaboration. This paper introduces an innovative approach that exploits the inherent natural language capabilities of pre-trained LLMs and VLMs to enable humans to interact with autonomous robots through natural language dialogues. As demonstrated in Figure \ref{fig:exampleResults}, we aim to realize a new approach to human-robot interactions—one where the conversation is the command (refer to Sections \ref{sec:3} \& \ref{sec:4} for more details).


Our contributions are therefore threefold: (a) we introduced a framework that can leverage independent pre-trained LLMs (e.g., OpenAI GPT-2 \cite{gpt2} \& GPT-3 \cite{gpt3}, Google BERT \cite{bert}, Meta AI LLaMA \cite{llama}, etc), and VLMs (e.g., CLIP\cite{clip}) to enable real-world autonomous robots to interact with humans or other entities using natural language dialogues. (b) we performed real-world experiments with our developed framework to ensure that the robot's actions are always aligned with the user's instructions, thereby reducing the likelihood of erroneous operations. (c) we have made our code and associated resources available to the public. This allows for easy reproducibility of our results.

\section{Related Work}
The recent rise of natural language processing (NLP) \cite{langNLP}, marked by large language models (LLMs) like  OpenAI GPT-3 \cite{gpt3}, Google BERT \cite{bert}, HuggingFace distilBERT \cite{distilBert}, EleutherAI GPTNeoX \cite{gptneoX}, Meta AI LLaMA \cite{llama}, Facebook RoBERTa \cite{roberta}, multi-modal vision-language models (VLMs) e.g., CLIP \cite{clip}, DALL-E \cite{dalle}, and their successors, has opened new avenues for human-robot interaction. The inherent capacity of these models to understand and generate human-like text as well as visual observations has led to several interesting applications \cite{brohan2023rt2}, \cite{lang1}. Recent works such as \cite{lang}, \cite{lang1}, \cite{lang2} and \cite{lang3} have successfully incorporated LLMs and VLMs into robotic systems, allowing the LLMs to interpret and execute complex commands. Similarly, Wangchunshu et al.\cite{zhou2023agents}, Kaiwen Zhou et al. \cite{esc}, and Miguel et al. \cite{llmROS} in their work demonstrated how LLMs could be used to facilitate real-time feedback, zero-shot object navigation and cognitive learning in autonomous robots.

While these works are exceptional, they have focused solely on a step-by-step task description \cite{lang3}, and rely completely on the LLM's ability to plan the robot's actions and act. In most complex real-world scenarios, especially as LLMs can sometimes hallucinate \cite{llmHallucination} or generate inconsistent data, their approach may introduce inconsistencies and randomness in the robot's actions.

On the contrary, our approach draws inspiration from the work of Yagi Xie et al. \cite{langNLP}. Instead of relying completely on the LLMs' ability to plan and execute the robot's actions, we employ a bidirectional approach, simply using the LLMs as a linguistic decoder \cite{langNLP}, and a classical robot operating system (ROS) \cite{ros} navigation planner to plan the robot's actions. We provide the LLMs with a dictionary of task descriptions and action patterns. We then use the ROS planner\footnote{\url{https://github.com/ros-planning/navigation}} to plan the actual physical actions of the robot (e.g., path planning, localisation, obstacle avoidance, mapping, etc.), as shown in Figure \ref{fig:exampleResults}b.

\section{Methods}\label{sec:3}
\begin{figure}[htp]
 \centering
  \includegraphics[scale=0.225]{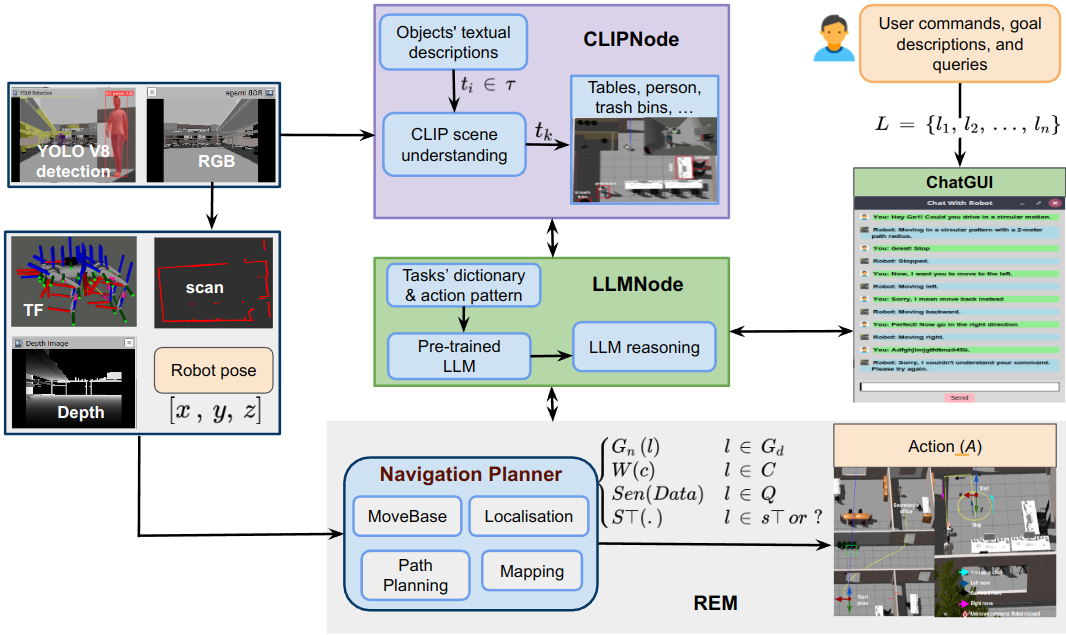}
 \caption{Overview of our framework's architecture. The LLMNode decodes the natural language conversations. The CLIPNode provides a visual and semantic understanding of the robot's task environment. The REM node abstracts the high-level understanding from the LLMNode to actual robot actions. The ChatGUI serves as the user's primary interaction point. See Subsections \ref{gpt}, \ref{rem}, and \ref{gui} for more details.}
    \label{fig:method}
    \Description{This figure is a flowchart describing our method's architecture. At the top left, a YOLO V8 detected and segmented object as well as an RGB image used as visual inputs is shown. This data flows into ``CLIPNode'', which understands scenes using textual descriptions such as ``Tables, persons, trash bins etc.''.
    Adjacent to this, on the right, there's a chat interface ``ChatGUI'' where a user inputs commands and receives robot responses. Below, in the centre, are images labelled ``TF (robot's transformation)'', ``Scan (lidar data)'', and ``Depth (depth image)'' used as inputs to the navigation planner.  There is also a block labelled ``Robot pose'' indicating the robot's spatial orientation and odometry data with \(x, y, z\) coordinates.
    In the centre is the ``LLMNode'' for command interpretation. Below is the ``Navigation Planner'' with elements for movement control (``MoveBase''), location estimation (``Localisation''), and route creation (``Path Planning'').
    Finally, on the bottom right, is the ``Action (A)'' section representing the robot's execution mechanism (REM) in the simulated environment layout. The entire architecture illustrates the integration of vision, large language models, and ROS planner for robot navigation.}
\end{figure}

An overview of our framework's architecture is shown in Figure \ref{fig:method}. One of our core objectives is to develop a framework that enables real-world autonomous robots to interact with humans or other entities using natural language dialogues. 
To achieve this objective, we decompose the task into three subtasks: (a) the integration of LLMs and VLMs, (b) the development of the robot execution mechanism (REM) node, and (c) the development of the chat graphical user interface (ChatGUI). An overview is shown in Figure \ref{fig:method}. This section provides details of each of the above three subtasks. 

\subsection{\textbf{Integration of LLMs and VLMs}}\label{gpt}
To decode the natural language conversation and abstract them to the robot's actions, we developed a ROS node, LLMNode (light green block in Figure \ref{fig:method}) to establish communication between the pre-trained LLMs and the rest interfaces within ROS ecosystem \cite{ros}. The LLMNode subscribes to topics that provide essential data, e.g., odometry for spatial sensing and outputs from the CLIPNode (light purple block in Figure \ref{fig:method}) for visual observation and object recognition. We used the LLMNode to handle incoming natural language inputs from the ChatGUI (Subsection \ref{gui}) by first passing them through the pre-trained LLM \cite{gpt2}. The output is then mapped to the robot's actionable commands or queries.

In the mapping process, we leverage pattern matching to align the generated text with predefined actions or information requests. For example, navigation commands are translated into goals for the robot to pursue within its environment, while queries \(\mathcal{Q}\) about the robot's status or surroundings are addressed with information derived from the robot's sensor data. The LLMNode also oversees the execution and feedback process.  We added this function to provide real-time feedback through the publishing of messages, which not only inform the user but also log the interaction data for subsequent analysis (see Subsection \ref{participant} for more details).

Summarily, the LLMNode function can be described as a mapping from natural language inputs to robot actions, i.e., \(\text{LLMNode}: \mathcal{L} \mapsto \mathcal{A}\) where \(\mathcal{L}\) represents the space of natural language inputs and \(\mathcal{A}\) denotes the set of possible robot actions. This mapping is a composition of several functions, as depicted in Eq. \ref{eqn:gpt}.
\begin{equation}\label{eqn:gpt}
    \text{LLMNode}(l) = \text{REM}(\text{LM}(l), \ \text{Sen}(\text{Data})), \  l \in \mathcal{L}
\end{equation}
From Eq. \ref{eqn:gpt}, \(\text{LM}(l)\) is the language model's interpretation of the input \(l_{i} \in \mathcal{L}\) and \(\text{Sen}(\text{Data})\) represents the sensor data that informs the context of the command. The \(\text{REM}\) node (Section \ref{rem}) then translates this into an executable command for the robot.

Furthermore, to provide a visual and semantic understanding of the task environment (e.g., Figure \ref{fig:exampleResults}b), we used the OpenAI contrastive language image pretraining (CLIP) model \cite{clip}. CLIP model consists of language and image encoders trained on a staggering 400 million image-text pairs \cite{clipData}. Thus, we used it to encode the stream of RGB images from our observation source (Intel Realsense D435i) alongside the textual descriptions of objects in the image.

Formally, given an image \(\mathcal{I}_t\) at time \(t\), we consider a set of predefined textual descriptions \(\mathcal{D} = \{d_{1}, d_{2}, ..., d_{n}\}\). Each description \(d_{i} \in \mathcal{D}\) is mapped to a tokenized representation, forming a set \(\mathcal{T} = \{t_{1}, t_{2}, ..., t_{n}\}\). This set encompasses human-readable labels for common office objects such as ``table'', ``chair'', ``person'', and so on. Using the CLIP model \cite{clip}, we extract the feature vector of the image, i.e., \(f_{\mathcal{I}} = \text{CLIP\_encode\_image}(\mathcal{I}_t)\). For each tokenized description \(t_i \in \mathcal{T}\), its feature vector is obtained as \( f_{\mathcal{T}_i} = \text{CLIP\_encode\_text}(t_i)\). Subsequently, for the image feature and every text feature, we compute the similarity scores using \(\mathcal{S}_i = f_\mathcal{I} \cdot f_{\mathcal{T}_i}^\mathcal{T}\).
Thus, we determine the recognized object within the image by selecting the textual description that yields the highest similarity score, as depicted in Eq. \ref{eqn:1}.
\begin{equation}\label{eqn:1}
\text{Recognized Object} = t_k ; \ k = \arg\max_{i=1}^{n} \mathcal{S}_i
\end{equation}

Additionally, our model uses the bounding boxes from YOLO V8 \cite{yolov8} to determine regions of interest (ROI) within the image. Notably, the centres of these bounding boxes are employed as the spatial coordinates for the recognized objects, capturing both the identity and the location of the objects in the scene. 

We embodied the entire process within a ROS node, CLIPNode (light purple block in Figure \ref{fig:method}). The output from Eq. \ref{eqn:1}, representing the recognized objects along with their respective spatial coordinates, is published as a ROS \cite{ros} topic. These are subsequently subscribed to by the LLMNode to handle the natural language commands, generate responses, and decide on actions for the robot to take. For instance, based on the prompt used by the robot's user, it can direct the robot to navigate to a detected object or provide information about detected objects and their positions.

\subsection{Robot Execution Mechanism (REM)}\label{rem}
To abstract the high-level language understanding and environment sensing from the LLMNode to actual robot actions, we developed the robot execution mechanism (REM) node. This node translates intents extracted from the LLMNode into actionable tasks for physical execution by the robot. Central to the REM node's functionality is processing navigation goals, \(\mathcal{G}_{n}\) (e.g., Figure \ref{fig:exampleResults}b). When a textual description of a goal destination \(\mathcal{G}_{d}\), such as ``navigate to the Secretary's office'' is provided, the REM node translates this into precise goal coordinates \((x_{l}, y_{l}, z_{l}, w_{l})\) within the robot's operational environment via a mapping process that correlates the descriptive labels with their corresponding spatial coordinates i.e., \(\mathcal{G}_{d} \mapsto (x_{l}, y_{l}, z_{l}, w_{l})\). To navigate the robot to the goal, we used the \textit{MoveBase} package of the ROS navigation planner, which provides an action server for handling navigation goals. REM node sends the goal to this server and monitors its progress.

Besides navigation goals, the REM node also handles custom movement commands \(\mathcal{C}\) (e.g., Figure \ref{fig:exampleResults}a) which are not tied to specific goal locations, but rather to particular motion patterns, such as ``rotate in place'', ``move forward'' etc. We encoded these patterns into the robot's YAML configuration files, allowing for a flexible command set \(c_{i} \in \mathcal{C}\) that can be expanded or modified as required. We use the REM node to translate the commands into \textbf{Twist} messages \(\mathcal{W}\) with linear \((v_{x},v_{y}, v_{z})\) and angular \((\omega_{x},\omega_{y}, \omega_{z})\) velocity components as \(\mathcal{W}(c) = Twist(v_{x},v_{y}, v_{z}, \omega_{x},\omega_{y}, \omega_{z})\).

In addition to handling movement, we integrated a security measure to halt the robot when an unrecognized command (e.g., the last command in Figure \ref{fig:exampleResults}a) is received, issuing zero velocities to stop all motion, ensuring safe operation.
\begin{equation}\label{eqn:navstack}
    \text{REM}_{l} = 
    \begin{cases}
    \mathcal{G}_{n}(l), & \text{if } l \in \mathcal{G}_{d} \\
    \mathcal{W}(c), &  \text{if } l \in \mathcal{C} \\
    \text{Sen(Data)}, &  \text{if } l \in \mathcal{Q} \\
    Stop(), &  \text{if } l \in \text{stop or unknown command}\\
    \end{cases}
\end{equation}
Formally, as summarised in Eq. \ref{eqn:navstack}, the REM node abstracts the complexity of the robot navigation and command execution, translating the high-level instructions into physical actions.

\subsection{Chat Interface Development}\label{gui}
To provide an intuitive conversational platform that would facilitate natural language interaction between the robot and its human users, we developed a simple and user-friendly chat graphical user interface (ChatGUI) which serves as the user's primary interaction point with the robot through textual communication. We designed the ChatGUI using Tkinter libraries\footnote{\url{https://docs.python.org/3/library/tkinter.html}} and integrated it within ROS \cite{ros} for message passing. 
We employed the standard ROS publish/subscribe communication mechanisms for the ChatGUI development, specifically, a bidirectional message exchange approach, i.e., \(\text{ChatGUI}: \text{UserInputs}  \leftrightarrow \text{LLMNodeOutputs}\). User natural language inputs are published to the LLMNode, and the responses are subscribed to and displayed to the user on the ChatGUI interface.

We developed the ChatGUI with event-driven architecture to ensure that user actions, such as sending a message or issuing a command, trigger corresponding updates in the ChatGUI or result in the publishing of commands to the LLMNode. We encapsulated this process in a function that translates user actions into corresponding LLMNode responses.

\section{Preliminary Results}\label{sec:4}
We conducted real-world and simulated experiments to demonstrate the applicability and adaptability of our framework. For simulation, we used the Unitree Go1 Gazebo packages\footnote{\url{https://github.com/unitreerobotics/unitree_guide}} 
and a ROS-based open-source mobile robot adapted from \cite{romr}. We ran all the simulations with a  ground station PC with Nvidia Geforce RTX 3060 Ti GPU, 8GB memory running Ubuntu 20.04, ROS Noetic distribution.

In real-world experiments, we used a Lenovo ThinkBook Intel Core i7 with Intel iRIS Graphics running  Ubuntu 20.04, ROS Noetic distribution. Unitree Go1 quadruped robot was used. The robot is equipped with Intel Realsense D435i RGB-D camera and \(\text{Ouster} ~3D \) LiDAR for both visual and spatial observations of the task environment. All the real-world experiments were performed in our laboratory office (11 rooms) and outside corridor environment, measuring approximately \(18 \times 20\;m\) and \(6 \times 120\;m\) respectively. 

We experimented with different pre-trained open-source LLMs like OpenAI GPT-2 \cite{gpt2}, Google BERT \cite{bert}, HuggingFace distilBERT \cite{distilBert}, EleutherAI GPTNeoX \cite{gptneoX}, Meta AI LLaMA \cite{llama}, and Facebook RoBERTa \cite{roberta}. OpenAI GPT-3 \cite{gpt3} and GPT-4 \cite{gpt4} are also adaptable to our framework. However, due to their API access limitations, we mostly utilised the open-access and free versions of the LLMs (GPT-2 \cite{gpt2} specifically) in our experiments. 

\subsection{Initial Evaluation / Participants}\label{participant}
\begin{figure}[htp]
    \centering
   \includegraphics[scale=0.18]{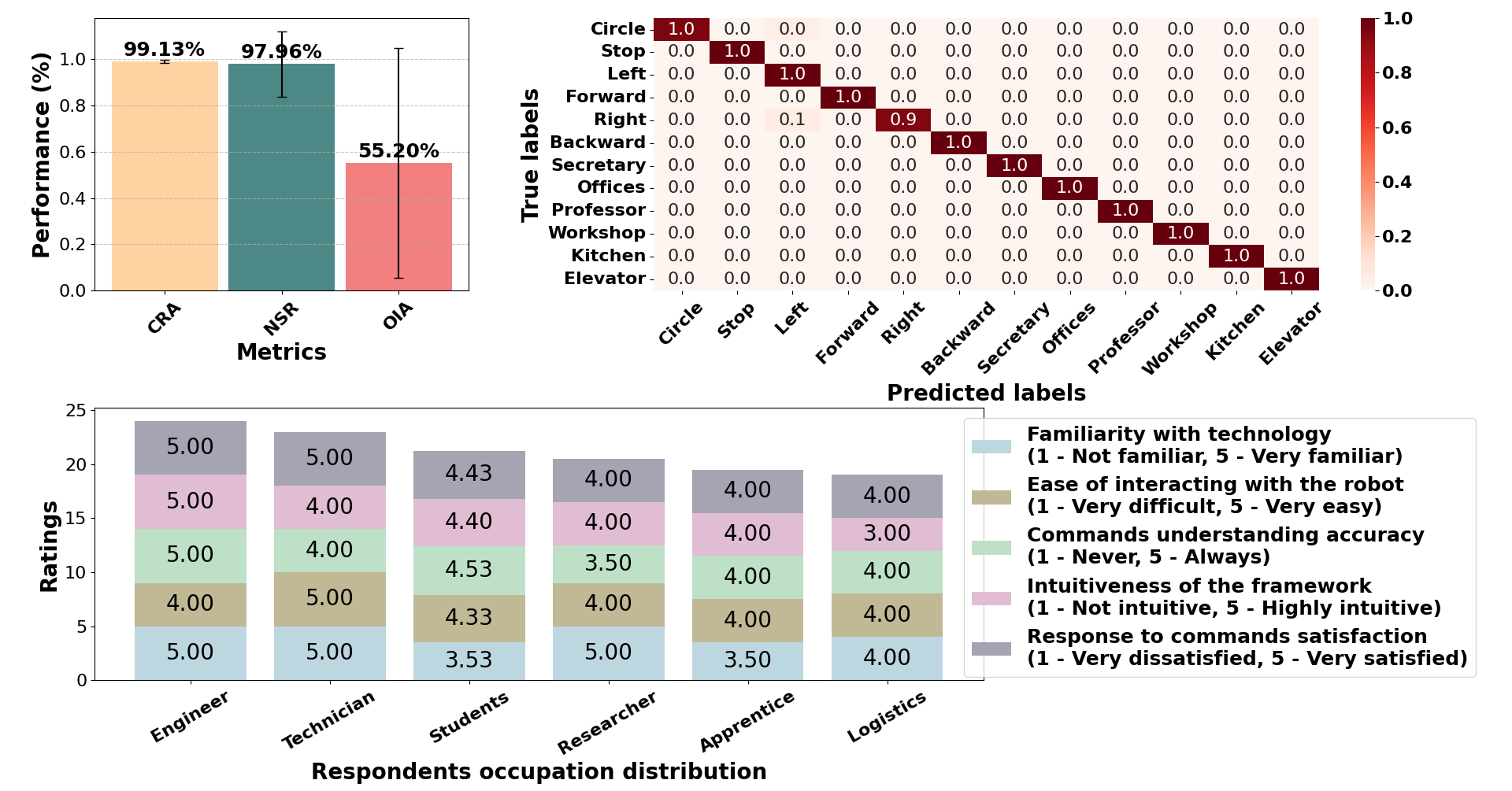}
    \caption{Performance and variability measures illustrating CRA, OIA, and NSR (top) and the participants' feedback (bottom) based on the logged interaction data.}
    \label{fig1}
    \Description{This figure shows a statistical analysis of interaction data logged from participants' interactions with the robot. At the top left is a bar chart with three different metrics for accessing the performance accuracy of our framework which includes the command recognition accuracy (CRA), object identification accuracy (OIA), and the navigation success rate (NSR). The CRA and NSR have very high-performance percentages, 99.13\% and 97.96\% respectively, whereas OIA has a significantly lower performance at 55.20\%. The top right features a confusion matrix table used to describe how the LLMNode predicted labels diverge from the true labels. Each cell in the matrix shows a number, with a scale from $0.0$ to $1.0$, where $1.0$ represents a perfect match between the predicted and the true label.
    The bottom row shows a stacked bar chart displaying ratings from participants from different occupation distributions like Engineer, Technician, Student, and others. Five aspects are rated on a scale from $1$ to $5$, such as familiarity with technology, ease of interacting with the robot, command understanding accuracy, intuitiveness of the framework, and satisfaction with the robot's response to commands. Engineers and Technicians gave the highest ratings across all aspects, while Students and Researchers varied, with some aspects rated as low as 3.5. Apprentices and Logistics participants gave middle-ground ratings, no lower than 3.5 and no higher than 4.5.}
\end{figure}

In our initial evaluation, we invited \(21\) participants (mostly students) with an average age of \(23 \ (\pm 5)\) and gender distribution, \(61.9\%\) male,  \(28.6\%\) female and \(9.5\%\) others to assess the intuitiveness of our framework by interacting with the robots using natural language. We instructed the participants to command the robots to navigate to locations, identify objects, and inquire about their status. We meticulously logged the interaction data which includes the participant's input text, the LLM's output, the true label, the LLMNode predicted label, etc. To quantitatively evaluate the performance of our framework, we established four key metrics:

    \textbf{(a) Command Recognition Accuracy (CRA)}: With the CRA, we assess how accurately the LLMNode interprets the natural language commands. This aids us in pinpointing instances where the predicted label diverged from the true label, providing insights into potential areas for improvement.
     \textbf{(b) Object Identification Accuracy (OIA)}: We employed this metric to assess the precision of the CLIPNode in identifying and localizing objects within the robot's task environment.
     \textbf{(c) Navigation Success Rate (NSR)}: We utilised this metric to determine the effectiveness of the REM node in successfully navigating the robot to the designated locations.
     \textbf{(d) Average Response Time (ART)}: We logged in ROS Unix epoch clock standard, the time a message is sent from the ChatGUI, the time it is received by the LLMNode, and the time the robot responds. With the ART, we compute the average duration from receiving the user's chat command to initiating the robot's movement. \\
Figure \ref{fig1} presents our preliminary statistical results obtained from the interaction data analysis. The top row of Figure \ref{fig1} shows the performance metrics and the confusion matrix (for selected labels) of the LLMNode. The CRA, with a prediction accuracy of \(99.13\%\) (i.e., how often the ``Predicted labels''  matched the ``True labels''), indicates a high level of accuracy in the command interpretation. This reflects the robustness of the LLMNode in processing the natural language inputs. The OIA on the other hand, achieved \(55.20\%\) accuracy, indicating room for improvement in our CLIPNode integration. Further, the NSR at \(97.96\%\), indicates good performance in the REM's ability to abstract the high-level understanding from the LLMNode to the actual robot's navigation actions.
Also, the overall ART across all the selected commands (refer to the figure at \url{https://osf.io/ufctx}) is approximately 0.45 seconds. This indicates that, on average, the robot takes less than half a second from receiving a chat command to initiating movement, which suggests a relatively quick response time for our framework.

Furthermore, the bottom row of Figure \ref{fig1} shows the participants' feedback (refer to the questionnaire at \url{https://osf.io/dgbtr}). With $4$ and $5$ ratings as favourable benchmarks, \(80.9\% \) and \(76.2\%\) of the participants respectively rated the ease of interaction and the intuitiveness of our framework as favourable, while \(85.7\% \) are satisfied with the response of the robot to their natural language commands.

\section{Conclusion and Future Work}
We introduced a framework that leverages the inherent capabilities of large language models (LLMs) and multimodal vision-language models (VLMs) to enhance human-robot interaction through natural conversation. Our evaluation from the logged interaction data and participants' feedback was overwhelmingly positive. The high command recognition accuracy and effective task execution, show that our framework can simplify human-robot interaction.
Looking ahead, we aim to refine the framework across several dimensions, not just for ROS-based autonomous robots.
The CLIPNode will be further improved for broader object recognition, and the LLMNode will be fine-tuned with domain-specific data for better contextual and voice understanding. User experience will be a priority, with a focus on creating a more intuitive and adaptive chat interface.

\bibliographystyle{ACM-Reference-Format}
\bibliography{references}

\end{document}